# Layers of technology in pluriversal design.

# Decolonising language technology with the LiveLanguage initiative

*Gertraud Koch[a], Gábor Bella[b], Paula Helm[c], Fausto Giunchiglia[d]*

[a] Department of Anthropological Studies in Culture and History, University of Hamburg, Hamburg, Germany; [b] Department of Logic of Uses, Social Sciences, and Information, IMT Atlantique, Nantes, France; [c] Department for Media Studies/Data Science Center, University of Amsterdam, Amsterdam, Netherlands; [d] Department of Information Engineering and Computer Science, University of Trento, Trento, Italy

Gertraud Koch: gertraud.koch@uni-hamburg.de; Gabor Bella: gabor.bella@imt-atlantique.fr; Paula Helm: p.m.helm@uva.nl; Fausto Giunchiglia fausto.giunchiglia@unitn.it

Language technology has the potential to facilitate intercultural communication through meaningful translations. However, the current state of language technology is deeply entangled with colonial knowledge due to path dependencies and neo-colonial tendencies in the global governance of artificial intelligence (AI). Language technology is a complex and emerging field that presents challenges for co-design interventions due to enfolding in assemblages of global scale and diverse sites and its knowledge intensity. This paper uses *LiveLanguage*, a lexical database, a set of services with particular emphasis on modelling language diversity and integrating small and minority languages, as an example to discuss and close the gap from pluriversal design theory to practice. By diversifying the concept of emerging technology, we can better approach language technology in global contexts. The paper presents a model comprising of five layers of technological activity. Each layer consists of specific practices and stakeholders, thus provides distinctive spaces for co-design interventions as mode of inquiry for de-linking, re-thinking and re-building language technology towards pluriversality. In that way, the paper contributes to reflecting the position of co-design in decolonising emergent technologies, and to integrating complex theoretical knowledge towards decoloniality into language technology design.

## 1. Introduction [1]

Language plays a crucial role both for (neo-)colonial power through reproducing the linguistic hierarchy on a global scale (Grosfoguel 2011, 10) and imperial knowledge cast in Western imperial languages (W. D. Mignolo 2009) and for creating a repertoire for delinking, relinking and rebuilding decolonised ways of living in a pluriverse (Escobar 2020; W. Mignolo and Walsh 2018). Diversifying existing language databases to better represent the linguistic diversity is an important pillar of decolonisation, as it allows for greater access and opportunities for self-expression for small language groups (Graham and Zook 2013) and meaningful translation for intercultural dialogue on transitions towards sustainable futures (Escobar 2018 2020). Nevertheless, the collection of data on minority languages carries the risk of new forms of epistemological colonisation (Spivak 2010), epistemic injustice (Fricker 2013), cultural appropriation (Benjamin 2019) or epistemicide (Santos 2007), when corporate and military interests or top-down approaches prevail (Helm, De Goetzen, et al. 2023).

The investigation into the decolonisation of language technology (LT) starts by outlining the capacity of language and its transition into LT to promote transitions to a pluriverse, a world in which many worlds fit (Escobar 2015; W. D. Mignolo 2009), while still entangled in the colonial triangle (W. Mignolo and Walsh 2018). When examining the colonial trajectories in LT, we investigate whether language diversity and plurality of meaning are concerns of value-oriented design (Donia and Shaw 2021) and co-design in the field of emergent technologies (Lanzeni and Pink 2021). As an example, we take LiveLanguage, a lexical database, catalogue and set of services around it, which surpasses existing language databases in terms of the number of languages represented and, more importantly, in the way it is organised. However, despite all the efforts towards value-oriented design, the concepts of diversity and plurality, referred to as diversality and pluriversality by Escobar (2020), are not yet fully implemented. The work on LiveLanguage highlights that value-oriented design alone is insufficient for decolonising technology in the social world.

Technological development and the design of language technologies, including large language models (LLMs), occur across multiple fields of technological activity. This encompasses a wide range of social practices that are relevant for co-design interventions. It is important to note that LLMs are a particular case of language technologies.

The concept of technological activities shows that plurality is a design issue, a requirement and a consideration on five permeable layers of technological activity in LT development. This disaggregation into multiple layers of technological design diversifies the entry points for interventions through participatory and co-design practices, thus suggests plurifying co-design itself (Smith et al. 2021), even though in a technical dimension.

## 2. The relevance of language for transitions towards pluriversality

Decolonial theory acknowledges the central role of language in constructing a Christian, Western, and European identity and implementing this geopolitics through a 'totalitarian and epistemically undemocratic implementation' of Western imperial knowledge worldwide (W. D. Mignolo 2009, 176). The use of Western imperial languages was prevalent in disseminating imperial knowledge. Myth, folklore, and traditional knowledge have been used to legitimise the imperial epistemology of the sciences with a formal apparatus of enunciation and description (W. D. Mignolo 2009, 177). These linguistic traditions continue to play a crucial role in maintaining colonial domination. Speaking Western languages is seen as a way of becoming 'whiter', and translation into major languages is essential for having a voice at all (W. D. Mignolo 2009, 165ff). Decolonial theory thus identifies the linguistic hierarchy as one among other power hierarchies on a global scale (Grosfoguel 2011, 10).

Against the background of this domination of Western ideologies embedded in knowledge and language and imposed on the Global South, decolonial theory puts forward the idea of overcoming abyssal thinking (Santos 2007) towards an ecology of knowledges. Rather than perpetuating the dominance of techno-scientific knowledge, decolonial theory argues for the recognition of the incompleteness of all knowledges and the recognition of the heterogeneity of knowledges as well as the persistent and dynamic interconnections (Santos 2007, 66). In this understanding, knowledge is an intervention in, not a representation of, the world (Santos 2007, 70). Taking into account the differences of knowledge and its exemplifications in languages is important to avoid further epistemicide, which has been going on for the last five centuries through techno-scientific interventions (Santos 2007, 74). Some proponents of decolonial theory advocate for a way of thinking that does not separate thinking from feeling and knowledge from care (Escobar 2020). At the same time, decolonial theory emphasises the crucial role of language in knowledge production as a common human endeavour that requires a semiotic code shared between users in semiotic exchange (W. D. Mignolo 2009, 176). Language, and the translation of languages, is a precondition for speaking, reflecting and discovering the diversity of knowledge and the value of knowledge ecologies. Translation is important for learning from the Global South (Santos 2007, 66). Therefore, the process of translation is crucial, and it is important to consider the translators' awareness of conceptual differences and potential blind spots to the situated knowledge of others.

> To recuperate some of these experiences [of epistemicide], the ecology of knowledges resorts to intercultural translation, its most postabyssal feature. Embedded in different Western and non-Western cultures, such experiences use not only different languages but also different categories, symbolic universes, and aspirations of a better life. (Santos 2007, 74)

Intercultural translation encounters the challenge of dealing with the 'incommensurability, incompatibility, and mutual unintelligibility of knowledge' (Santos 2007, 75). Despite this, it remains a crucial tool for expressing, perceiving, and exchanging ideas among people with diverse linguistic backgrounds (Santos 2007, 74). Diversity-aware translations aim to bring together different ontologies and value the use of poetic, artistic, spiritual, and subjective methods of knowledge production (Anzaldúa 2015). This approach is crucial for transitioning towards a pluriverse, where multiple worlds can coexist. As Tlostanova (2017, 54) suggests, this requires a shift away from Western-centric concepts and towards a recognition of non-Western parallels and opposites.

### 3. Language technology – colonial trajectories

Given the complex relationship between language and technology in the colonial triangle, questions arise about the role of LT in facilitating appropriate translations across all languages and dialects for intercultural dialogues. A crucial challenge in this sense is how to de-link the dominant entanglements of LT in the colonial triangle, to then re-think and re-build them towards the ideas of decolonial theory on linguistic diversity, diversity-aware translation and discourses of transition. Delinking from colonial knowledge traditions is necessary in several aspects.

*Path dependency of language technology*

The history of LT goes back to the 1960s, when the first English dictionaries went online, which later became an important source for training AI machines. Digital dictionaries were based on the semantic structure of a tree graph, the Princeton standard, created by humans; European projects followed. English graphs were attached to other languages, making English a 'hub language'. This involved using the existing graph as it was and relating it to the other languages, resulting in lexical gaps and quality problems in translations due to differences in vocabulary (e.g. Pianta et al. 2002). From these initial language resources, several improvements and innovations followed, such as going beyond words by integrating concepts into the dictionaries (e.g. Giunchiglia et al. 2010). When India developed its own lexicon with Hindi as the hub language (e.g. Redkar et al. 2017), a new path beyond the Princeton standard of Anglo-European languages opened up, followed by a new option of combining the hub language English with the hub language Indian, taking into account the additional concepts in each language (Giunchiglia, Batsuren, and Bella 2017). Still, for minority languages, the lack of adequate linguistic resources hinders their representation in LLMs. Furthermore, the shift from semantic graph structures of dictionaries to machine learning as the core principle of AI-based LT raises new issues. Machine learning is based on learning from existing language digitalisation and many other types of digital resources that are no longer developed by humans according to semantic criteria; machine learning misses language change and variation, as well as nuances in meaning, but stereotypes language into average meaning (Giunchiglia, Batsuren, and Bella 2017). Moreover, the linguistic databases and corpora on which the most powerful AI models are currently trained, and on which large digital infrastructures are built, are dominated by the languages of the colonial imperial powers, reproducing imperial knowledge cast in Western imperial languages (Helm, Bella et al. 2023; W. D. Mignolo 2009), perpetuating sexist, racist and capitalist biases and linguistic hierarchies on a global scale (Grosfoguel 2011, 10). As an inherent logic of LT (and LLM in particular), this even works against initiatives that promote LT for all.[2]

*Global governance of language technology*

Diversifying existing databases to better represent linguistic diversity is an important pillar of decolonisation, as it facilitates access, opportunities for self-representation for small language groups (Graham and Zook 2013), and intercultural dialogue. However, the collection of minority language data is mostly in the hands of large internet companies, sharing the problems of AI governance that reproduce colonial power relations in multiple ways. Gurumurti and Shami (20120199, 9) problematise the silencing of problems of AI governance in international collaborations with the Global South, highlighting four dimensions that are currently not taken into account, thus reinforcing neocolonial tendencies: (1) the erosion of autonomy of individuals and communities through the enclosure of both data resources and intelligence capital by powerful corporations, (2) the importance of specific intelligence advantages of communities and countries for economic self-determination in the AI paradigm, (3) economic surveillance and neo-imperialist control of critical AI infrastructure, and (4) the importance of global communication commons that need to be protected by international binding frameworks to ensure democratic principles in the automated public sphere by defining the obligations of state and non-state actors.

These silences in the governance of AI are again critical for the development of LT, they delineate the spectrum of problems that call for interventions towards decolonisation. There is no authority or institutionalised mechanisms to legitimise machine-readable language and translation, such as the national language boards or community committees for written language. The exploitation of language resources is largely in the hands of large internet companies. They use crowdsourcing approaches to incorporate small languages and First Nation languages into their Anglo-centric LT, offer them as open source for free exploitation by anyone (Butryna 2019), and exploit them themselves.[3]

### 4. De-linking language technology from colonial trajectories

De-linking 'implies working at the margins, at the boundary between hegemonic and dominant forms of knowledge, economic practices, political demands' (W. D. Mignolo 2007, 160), which in the case of language technologies – as in all areas of emerging technologies – faces multiple complexities, which also occur across scales, places and practices (Wahlberg 2022). This chapter responds to this problem of co-design by broadening the understanding of how emerging technologies gain political and technological facticity in global assemblages. It broadens the concept of technology itself by drawing attention to the methodological approach of 'technical' and 'technological activities', which offers specific starting points, equally, for anthropological research on and disengagement from colonial knowledge trajectories.

*The emergence of political and technological facticity of language technology*

Despite the long colonial trajectories, language technology is still an emergent technology due to the ongoing dynamic innovation in information technologies. It forms 'hybrid collectives of people and technology' (Callon 2004) that are multiple and diverse. The implementation of language technologies, as for emergent technologies in general, goes hand in hand with the transformation of the technology and continuous redesigns that create new variations of the technology (Callon 2004; Lanzeni and Pink 2021).[4] The political and technological facticity of language technologies emerges in 'reciprocal and discursive interaction with technological activities' (Pfaffenberger 1992, 282). Their specific functionality, meaning and uses in social contexts are subject to negotiation. Differing motivations to reshape the social distribution of wealth, power or status collide in what Brian Pfaffenberger, anthropologist of technology, calls 'technological dramas' as emerging technologies are implemented. They create legitimations to justify emerging technologies through myths (ideas) about the usefulness of the technology, the social contexts of its use, and rituals around the handling of the artefacts (Pfaffenberger 1992; Butot and van Zoonen 2022). The envisioned technological uses, yet to be realised, negotiate ideas about potential technological and social futures and are situated in normative frameworks or ecologies of knowledge (W. D. Mignolo 2009). The political and technological facticity of language technologies is a highly complex and ambiguous interplay of multiple actors, motivations, powers, situations and technological activities in a variety of globally distributed social fields. The activities that give rise to the technology are distributed in a global assemblage of actors, practices, infrastructures and regulations (Ong and Collier 2005), which in turn complicates the endeavour of co-design interventions towards pluriversality. Sorting and systematising these complexities in the emergence of language technology thus promises the possibility of differentiated and specific interventions into the social practices that contribute to the rise of the political and technological facticity of language technologies in global contexts.

*Assembling language technologies through technological activities*

These considerations and colonial trajectories suggest that LTs, and LLMs in particular, form as a social fact in assemblages of distributed agency in diverse bundles of practices. The fluidity and complexity of the assemblages make the whole difficult to grasp (Ong and Collier 2005) in research, and for design interventions it is still possible to trace formative elements through the application of specific lenses. Coupaye (2022) suggests opening the black box of technology by enriching ethnographic observations with the methodological concept of chaînes opératoires to deconstruct social actions (such as cooking a dish or designing a technology) into individual technical activities that contribute to the emergence of this social action. In fact, he locates this methodological approach at the micro level of observing chaînes opératoires of individual action, which is most relevant for studying the use of language technologies in contexts. The micro-perspective is not appropriate for studying or intervening in the knowledge production of LT laboratories that reproduce the path-dependencies of language technology, or of companies and state actors that silence injustices in governance approaches. In these fields, the observation of chaînes opératoires needs to be scaled up to the meso level. Rather than the individual enactment of social practices, the focus of research and co-design intervention here is on the chaînes opératoires of bundles of social practices that shape language technologies, such as the perpetuation of path dependencies in technological design, the silencing of neo-colonial tendencies in AI governance, or the negotiation of political and technological facticity in specific contexts of use.

By bringing the lens of technological activity (Coupaye 2022) to bear on the colonial trajectories of language technology, we have identified five bundles or sets of practices in this global assemblage of language technology (Wahlberg 2022) that are crucial for the emergence of the political and technological facticity of technologies (Pfaffenberger 1992) throughout the various processes of their implementation (Callon 2004; Lanzeni and Pink 2021). They thus provide starting points for decoupling from existing knowledge (W. D. Mignolo 2009), for rethinking the concepts and myths of use (Pfaffenberger 1992; Santos 2007), for reconfiguring the distribution of wealth and social status, and for opening up imagined technological futures to ideas of pluriversality (Escobar 2015). These formative sets of practices of technological activity, which we will outline in more detail in the next chapter, are:

(1) the technological core consisting of enabling software language technologies;
(2) the knowledge domain modelled in information systems, i.e. the languages exploited by the language technology, as from item (1);
(3) the diverse knowledge ecologies, i.e. the frictions of diverse ideas, myths and meanings, including the norms codified in legal and ethical frameworks;
(4) the political economies necessary to create and maintain the technology, and finally
(5) the contexts of use, which are most relevant to the broader social relevance of a technology.

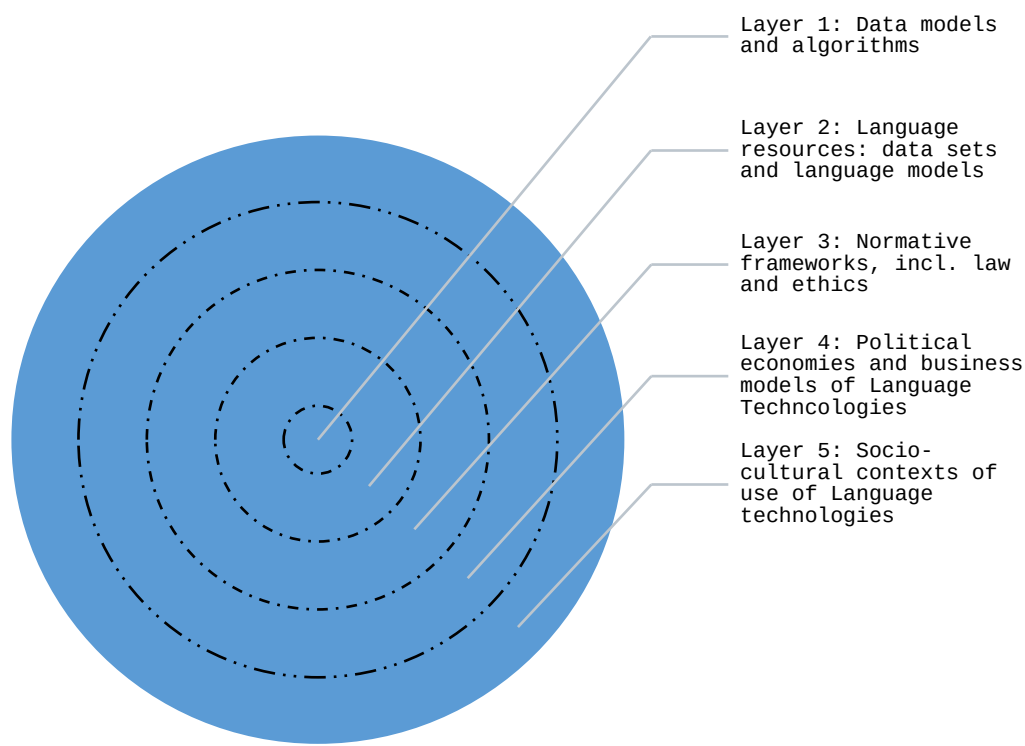

Figure 2. Layers of technological activity (Model by Gertraud Koch) [5]

In our model (Figure 2), we arrange these five sets of relevant practices of technological activities crucial to the political and technological facticity of language technology into permeable, interdependent layers that wrap around hardware and software technology, understood here as a material entity with its own agency (Callon 2004). Each layer contributes to the larger whole for the implementation of a technology and is specific in terms of actors, motivations, knowledge and materiality. Thinking in terms of permeable layers of practice bundles expands, diversifies and specifies the field of action for co-design approaches to pluriversal language technology without losing sight of the bigger picture. Each layer can facilitate or hinder disengagement from colonial trajectories, and malignant layers can cause serious problems, even expand and give the whole project a neo-colonial stance.

### 5. Re-thinking technological activities of language technology

Transformative practices in information technology towards decoloniality are still explorations, 'more mode than method; more tactic than strategy; more a way of proceeding than a field of object' (Philip, Irani, and Dourish 2012, 23). In order to discuss how the layers of technological activity can facilitate co-design approaches to decolonising language technologies and rethinking and rebuilding them towards pluriversal thinking, the chapter will use the example of LiveLanguage to outline, exemplarily for each layer, the bundle of practices of de-linking, rethinking and rebuilding towards diversality and pluriversality. The activities of LiveLanguage in this sense are, however, incomplete, experimental, unevenly realised in each layer, and they are not the result of co-design approaches to pluriversality, but of a Western discourse of transition (Escobar 2015, 20), motivated by dissatisfaction with the quality of AI-based translations and an appreciation of cultural diversity. Nevertheless, LiveLanguage is an example of the place of co-design in emerging language technology. The LiveLanguage initiative is built around the Universal Knowledge Core (UKC),[6] which is a multilingual lexical resource (layer 1) with a particular focus on linguistic diversity and cross-lingual lexical semantics, and provides services around it. The production of the UKC datasets has involved partners from all over the world in several projects.[7]

### 5.1. First layer: data models and algorithm

Data models and algorithms are one of the colonial trajectories of language technology outlined above. On a technical level, LiveLanguage shares this colonial trajectory with other European language technologies that use the Anglo-centric Princeton WordNet format, but has now undertaken several actions to break away from this trajectory by rethinking and rebuilding the data models and algorithms. A major activity is the work on lexical gaps in translations, which has been carried out in several steps over several years and goes hand in hand with the work in the second layer on language datasets. The emergence of the hub language Indian has resulted in the Universal Knowledge Core, a lexicon with a tree structure not based on words but on concepts, which can detect lexical gaps or indicate multiple meanings of words. This allows knowledge about the words and their meanings to be transformed into machine-readable form in interoperable formats for machine learning and natural language processing (Bella et al. 2022). At the level of data models, diversality (Escobar 2020) is a universal principle of the data model and algorithms in the UKC, albeit with further potential for improvement as work continues, for example on metonymy as a global form (Khishigsuren et al. 2022).

### 5.2. Second layer: language data sets and language models

While LLMs reproduce meanings in their learning materials, the design paradigm of UKC can de-link from colonial trajectories through technical work on the database language datasets and models, which are a scarce source for small languages and dialects (Cieri and Liberman 2019) and crucial for decolonising LT. Despite the fact most existing language databases and infrastructures only map the 40 most well-resourced languages, *UKC* is covering more than 2,000 lexicons already to varying degrees of completeness (Bella et al. 2022). Each new language, when developed, is natively translated directly with the other languages (Giunchiglia, Batsuren, and Freihat 2018), a unique feature, which applies to all languages and is a step towards plurifying the dominant Anglo-centric design. Diversity work in this layer often relies on finding available machine readable language resources and matching them with people with relevant language expertise. It is time-consuming and laborious, requiring manual data work to integrate languages at a conceptual level (e.g. Batsuren et al. 2019).

Each new lexicon, once integrated into the UKC, is by design translated (at the word and concept level) into the thousands of languages already included. On a technical level, the project consists of four steps: (1) Continuously collecting and storing data on languages; (2) monitoring their evolution, the goal of it being to study synchronic properties of the languages and to integrate them directly with each other, thereby avoiding pivoting around English, (3) using the datasets and the models derived from them for real-life applications, and (4) extending the UKC and LiveLanguage data and language models as needed by new needs (e.g. new applications, unforeseen features in new languages). Collaborative approaches with partners from different countries are in progress through cooperation with universities world wide. They aim to allow each language to develop its own lexicons, meanings and definitions, unconstrained by what can be expressed in other languages, for different temporalities of development in doing so, and for communication about meanings across languages and the creation of new hub languages, such as Arabic (Giunchiglia et al. 2021).

### *5.3. Third layer: normative frameworks*

While de-linking from colonial trajectories and rethinking have been ongoing activities in layers one and two for some time, the following layers have recently gained attention through the idea to work on diversity through service provision in LiveLanguage. In the first two layers, value-oriented design (Hendry, Friedman, and Ballard 2021) has proven to be an effective way of decoupling, rethinking and rebuilding LLMs through expertise and creativity engineering combined with a high motivation to push the quality of LLMs towards the conceptual diversity of LiveLanguage. The technological activities in the other layers do not directly require engineering skills, but indirectly refer to them and still shape engineering through meaningful and distributed practices and provide a broad space for co-design approaches.

In the third layer, technological activities circulate through non-technological epistemic cultures and specific knowledge domains that order and normalise the place of language technology in diverse knowledge ecologies (W. D. Mignolo 2009) through normative frameworks, including law and ethics. The diversity-oriented design of LiveLanguage in layer 1 and 2 has its limits where these regulatory or other normative frameworks set different emphases, and where diversity cannot be as easily defined as in LT as a variety of language concepts, but is contested, problematised, discredited or debated, and is a political issue in itself. Not everyone recognises diversity as a value in the same way, either because the term has been used for abyssal thinking or because of different, diversity-averse worldviews, or because they prefer the efficiency of language technology design. Pluriversal design theory offers heuristics for reframing such colonial trajectories, intervening in their agencies, and envisioning alternative futures (Mainsah and Morrison 2014; Schultz et al. 2018; Smith et al. 2020), which have yet to be explored in their potential for decolonising LLMs.

*5.4. Fourth layer: political economies and business models of language technology*

Another, perhaps even heavier, burden for establishing diversality in LT thus work towards pluriversality are the practices of global governance of AI and the silencing of the neo-colonial exploitation of knowledge and data resources at the expense of the Global South, outlined above as a colonial trajectory with well-documented problems of ownership, extraction and remuneration (Gurumurti and Shami 2019). At this stage, LiveLanguage seeks to circumvent these problems through coordinated efforts with and for language communities in an arena where, at the macro level, internet companies and states negotiate their interests with a high degree of legal and economic expertise and power. Recent research perspectives also link care, knowledge and agency for sustainable resource management with notions of stewardship (Enqvist et al. 2018). The expansion of these ideas into computing and data science is just beginning; they work towards trusted, cooperative, and commonly managed goods in the stewardship or ownership of communities and cultural groups (Hafen 2019; O'hara 2019; Pentland and Hardjono 2019; Tzouganatou 20232023; Zillner et al. 20212021). They open up a rich field for pluriversal co-design approaches that work against the neo-colonial political economies interwoven with layer five, the technological activities in the contexts and practices of LLM use.

## 5.5. *Fifth layer: contexts and practices of use*

The contexts of use play a crucial role for the emergence political and technological facticity (Callon 2004; Lanzeni and Pink 2021; Pfaffenberger 1992). They are thus an important field of activity in co-design, where the problems and pitfalls of the political economy of AI technology become manifest. The relevance of socio-spatial dimensions to technological processes is an important issue in co-design and the interventions into the paradoxes of the often neo-colonial character of technological transfer and innovation. They emphasise the right and need to join LT infrastructures as an important way of accessing resources and participating in the Internet (Calderón Salazar and Huybrechts 2020; Tachtler et al. 20212021). A wide range of issues, interventions and methods for involving local communities in a decolonial way in the design of new and information technologies are emerging in the co-design literature (Keskinen et al. 2022; Muashekele, Winschiers-Theophilus, and Koch Kapuire 2019; Smith et al. 2020; Stichel et al. 2019; Jahn and Segal 2023). LiveLanguage started working with local communities at the university level and aims for applications that are locally connected to the knowledge and language of non-academic communities, but co-design approaches to pluralism have yet to be explored for LiveLanguage and require inter- and transdisciplinary collaborations across sites.

## 6. Re-building: conclusions on the position of co-design in the transition of language technology towards decoloniality

Meaningful intercultural translations are most relevant for decolonisation (Santos) through dialogues of transitions towards the pluriverse (Escobar 2020), which gives LT a crucial role and responsibility for acknowledging diversity in language and knowledge. In order to decouple from colonial trajectories and to rethink and rebuild (W. D. Mignolo 2009) LT in co-design practice, the paper proposes to work on the gap between pluriversal design theory and practice by opening the black box of technology (Coupaye 2022), thus multiplying and diversifying the starting points for interventions in co-design practice. Thus, on a theoretical level, it extends the notion of emerging technologies in co-design to layers of technological activity in the global assemblage of LT, i.e. practices clustered around specific tasks and actors at different sites and scales (Wahlberg 2022), through which technology acquires its social facticity (Pfaffenberger 1992).

The disaggregation of the emergence of LT in bundles of practices invites to plurify pluriversal design approaches to LT according to the five different levels of technological activity, which can be seen as another mode of plurifying LT itself (Smith et al. 2021). Using the LT enabler LiveLanguage as an example, the paper discusses the position of co-design approaches for decolonising and plurifying language technologies. In the first two layers of (1) enabling software technology, and (2) language model and data, value-oriented design facilitated the integration of complex theoretical knowledge, which is a relevant and still challenging extension in co-design (Hurley, Dietrich, and Rundle-Thiele 2021; Mainsah and Morrison 2014; Muashekele, Winschiers-Theophilus, and Koch Kapuire 2019; Cruz 2021).

Layer (3), practices around the incorporation of LT into normative frameworks, and layer (5), contexts of use of language technologies, provide wide spaces for co-design interventions, though they are yet to be applied to LiveLanguage. In layer (4), co-design can more easily influence governance approaches to language technologies through activities in layers 3 and 5, as the macro-level of political economies of language technologies dominates layer 4 and is difficult to approach through interventions. Each layer provides and demands specific entry points for co-design intervention towards pluriversality.

LLMs are experiencing exponential growth and are having a significant impact on society. In this situation, there is a gap between theory and practice in co-design research, and there is a lack of research examples in this field. The agenda outlined in this paper, at the intersection of co-design and anthropology of technology, suggests potential areas of engagement. It aims to encourage research and the co-design community to explore LLMs, opening up new possibilities and broadening technological futures.

**Notes**

1. We thank the reviewers for their valuable feedback, which was both critical and supportive. The advice and questions we received prompted productive reflections on the empirical and theoretical aspects of decolonisation, co-design, and participation. They were greatly helpful in developing our argument.
2. https://lt4all.elra.info/proceedings/lt4all2019/2019.lt4all-1.0.pdf
3. In the initiative 'Language Technology for All' such collaborations are supported by the UNESCO, https://lt4all.org/en/index.html, 24.02.2024.
4. See exemplary Winschiers-Theophilus et al. (2019).
5. Model by Gertraud Koch.
6. http://ukc.datascientia.eu/concept
7.
For a detailed information see https://datascientiafoundation.github.io/LiveLanguage/